\documentclass[11pt]{article}

\usepackage[margin=1in]{geometry}
\usepackage[utf8]{inputenc} 
\usepackage[T1]{fontenc}    
\usepackage{url}            
\usepackage[numbers,sort&compress]{natbib}
\usepackage{booktabs}       
\usepackage{amsfonts}       
\usepackage{nicefrac}       
\usepackage[protrusion=true,expansion=false]{microtype} 
\usepackage[table]{xcolor}  
\newcommand{\rev}[1]{#1}
\usepackage{amsmath}
\usepackage{amssymb}
\usepackage{graphicx}
\usepackage{algorithm}
\usepackage{algorithmic}
\usepackage{multirow}
\usepackage{subcaption}
\usepackage{wrapfig}
\usepackage{hyperref}       
\usepackage{authblk}

\title{AffectVerse: Emotional World Models for Multimodal Affective Computing}

\author[1,2]{Bo Zhao\textsuperscript{*}}
\author[2]{Fanghua Ye\textsuperscript{*}}
\author[2]{Yixin Ji}
\author[3]{Sicheng Zhao}
\author[4]{Xiaojiang Peng}
\author[1]{Zitong Yu\textsuperscript{\dag}}

\affil[1]{Great Bay University}
\affil[2]{Tencent}
\affil[3]{Tsinghua University}
\affil[4]{Shenzhen Technology University}

\date{}

\begin{document}

\maketitle

\begingroup
\renewcommand\thefootnote{*}
\footnotetext{Equal contribution.}
\renewcommand\thefootnote{\dag}
\footnotetext{Corresponding author.}
\endgroup

\begin{abstract}
Humans infer emotions by integrating observed multimodal cues with expectations about how affective states may unfold. Existing multimodal large language models (MLLMs), however, often treat emotion recognition as static fusion over complete audiovisual-text inputs, leaving affective dynamics implicit. We propose AffectVerse, a Qwen2.5-Omni-based model equipped with an Emotion World Module (EWM), an action-free representation-level module for short-horizon latent affective prediction. \rev{EWM contains three modules: 1) Cross-Modal Temporal Imagination predicts future video/audio representations from past tokens with multi-step rollout. 2)  MAMA(Modality-Aware Multi-step Attention) Belief Aggregation compresses imagined tokens into modality-aware belief tokens. 3) Belief Injection inserts these belief tokens into the LLM for affective reasoning.} AffectVerse uses future prediction as a past-conditioned self-supervised signal: it does not replace modeling observed history or require unseen signals at inference, but forces the current belief state to encode transition cues that are predictive of subsequent affective change. Across nine benchmarks, AffectVerse improves at least 2.57\% over other models, while controlled ablations show additive gains from temporal imagination, cross-modal rollout, and belief aggregation. These results suggest predictive belief-state modeling is a practical alternative for affective computing.
\end{abstract}

\section{Introduction}
\label{sec:intro}

Consider a job interview in which a candidate's voice begins to quiver almost imperceptibly and their smile tightens into a mask-like expression. A human observer does not wait for the full utterance to conclude---they update a belief state over how the candidate's affect may evolve. This form of \emph{predictive emotional inference} is well documented in affective science~\cite{friston2010free,barrett2017theory}, yet current MLLMs often treat emotion recognition as static pattern matching over a complete clip.

\begin{figure}[t]
  \centering
  \includegraphics[width=\textwidth]{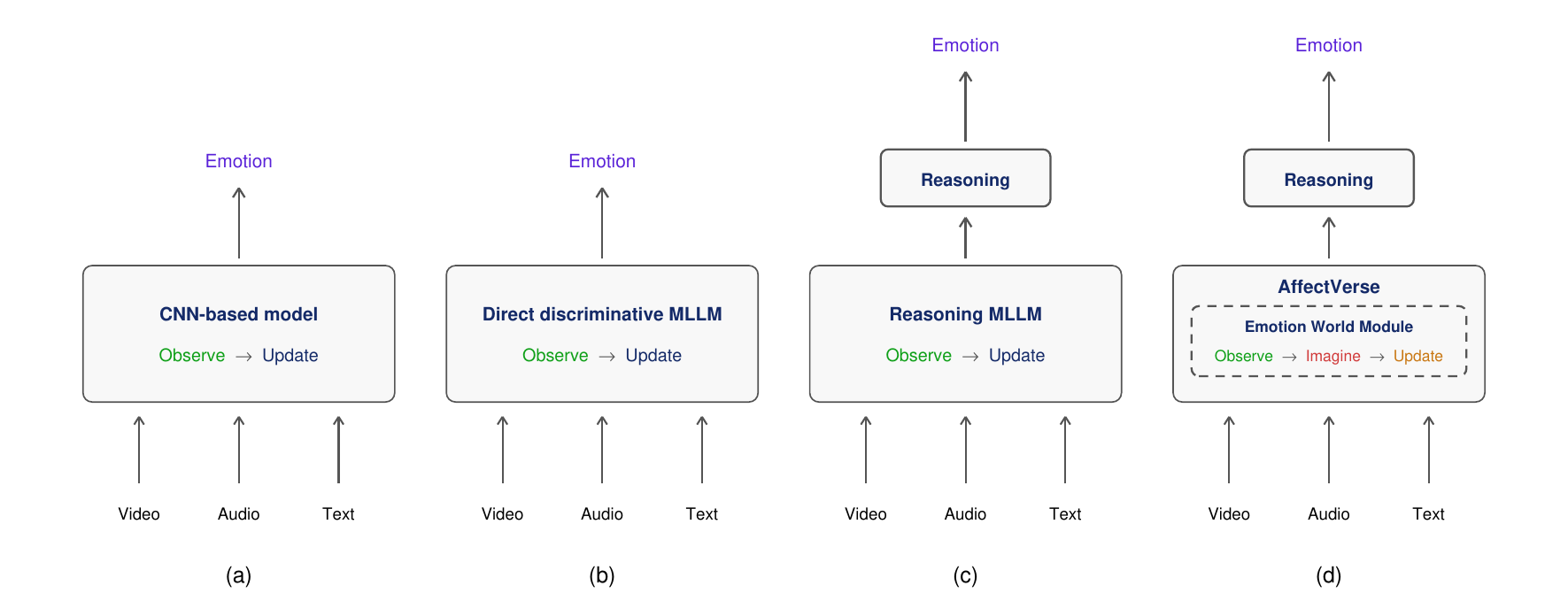}
  \vspace{-1em}
  \caption{\textbf{\rev{Motivation and positioning of AffectVerse.}}
\rev{AffectVerse introduces an Emotion World Module that inserts an intermediate Imagine stage, enabling the model to predict latent affective dynamics before updating the LLM's emotional context.}}
  \label{fig:motivation}
  \vspace{-1em}
\end{figure}

As illustrated in Figure~\ref{fig:motivation}, the method is motivated by the gap between static multimodal recognition and temporally unfolding affect. Existing MLLM-based methods~\cite{lian2023affectgpt,xu2024mplug,yang2024emovit} project multimodal features into a language model's embedding space and rely on the LLM's general reasoning ability to classify emotions. While effective, this paradigm often reduces affective dynamics---such as escalation, recovery, or cross-modal conflict---to a static snapshot. Future prediction is not meant to replace past-context modeling; instead, it provides a past-conditioned objective that encourages the representation to encode cues predictive of subsequent affective change.

 {Our key insight is that emotion understanding should not be formulated merely as static recognition over a complete input, but as predictive belief-state modeling under partial observations.} World-model-style predictive representation learning has been effective in reinforcement learning~\cite{ha2018world,hafner2021dreamerv2} and robotics, but affective inference requires an action-free formulation without rewards, policies, or control planning.
 {AffectVerse instantiates a representation-level variant for affective inference: it predicts latent future audiovisual representations from observed prefixes and uses the learned transition regularities to update the LLM's belief state.}

We present \textbf{AffectVerse}, whose EWM(Emotion World Module) instantiates the three main blocks in Figure~\ref{fig:framework}. Each block is tied to a distinct principle from affective science 
 {and motivated by converging evidence that emotion is not a fixed perceptual category, but a temporally unfolding process shaped by predictive inference, appraisal dynamics, bodily expression, multisensory evidence, and contextual interpretation~\cite{friston2010free,barrett2017theory,scherer2009emotions,ernst2002humans}}
:
\begin{enumerate}
    \vspace{-0.5em}
    \item \textbf{\rev{Cross-Modal Temporal Imagination.}} Grounded in predictive processing~\cite{friston2010free,clark2013whatever,barrett2017theory} and multisensory integration~\cite{ernst2002humans}, EWM predicts future video/audio representations from past tokens with cross-modal context, learnable future queries, multi-step rollout, and modality dropout.
    \item \textbf{\rev{MAMA(Modality-Aware Multi-step Attention) Belief Aggregation.}} Inspired by Scherer's Component Process Model~\cite{scherer2009emotions}, MAMA aggregates imagined video/audio tokens into compact belief tokens using type embeddings, dynamic belief queries, and boundary-state residual injection.
    \item \textbf{\rev{Belief Injection.}} The belief tokens are projected back to the LLM hidden space and inserted at two positions in the token sequence, allowing Qwen2.5-Omni to reason over both observed evidence and imagined affective continuations.
\end{enumerate}

\vspace{-1em}
\section{Related Work}
\label{sec:related}
\vspace{-0.5em}
\subsection{Multimodal Emotion Recognition}
\vspace{-0.5em}
Multimodal emotion recognition (MER) identifies human emotional states by jointly leveraging visual, acoustic, and linguistic cues~\cite{poria2019meld,zadeh2018mosei}. The field has progressed from modality-specific encoders with hand-crafted fusion~\cite{zadeh2016mosi} to attention-based fusion strategies and, more recently, MLLM-based approaches. Methods such as AffectGPT~\cite{lian2023affectgpt}, mPLUG-Owl~\cite{xu2024mplug},  {AffectGPT-R1~\cite{lian2025affectgptr1}, and Emotion-LLaMA v2~\cite{peng2026emotionllamav2}} leverage LLMs as unified multimodal reasoners, while  {recent fusion-oriented work explores modality complementarity~\cite{huang2026complementarity}.}
However, these methods primarily focus on feature alignment or fusion, without explicitly learning how emotional states evolve over time.  {AffectVerse instead equips the MLLM with an action-free predictive affective module that predicts latent affective transitions and constructs evolving belief states from partial observations.}

\vspace{-1em}
\subsection{World Models}
\vspace{-0.5em}
World models, originating from model-based reinforcement learning, learn internal representations of environment dynamics for future state prediction and planning. From compressed latent representations~\cite{ha2018world} to Dreamer V2/V3~\cite{hafner2021dreamerv2,hafner2023dreamerv3}, imagination-based training has shown strong potential in learning predictive dynamics for video prediction, autonomous driving, and embodied AI. The core appeal lies in predictive representation learning: instead of only reacting to current observations, a model learns latent transition structure that supports more robust downstream reasoning.  {EWM is inspired by this world-model-style predictive representation learning, but it is not a control-oriented world model: it does not model actions, rewards, policies, or planning. Instead, AffectVerse uses an action-free module to predict short-horizon latent audiovisual continuations and convert them into belief tokens for emotion interpretation.}

\begin{figure}[t]
  \centering
  \includegraphics[width=\textwidth]{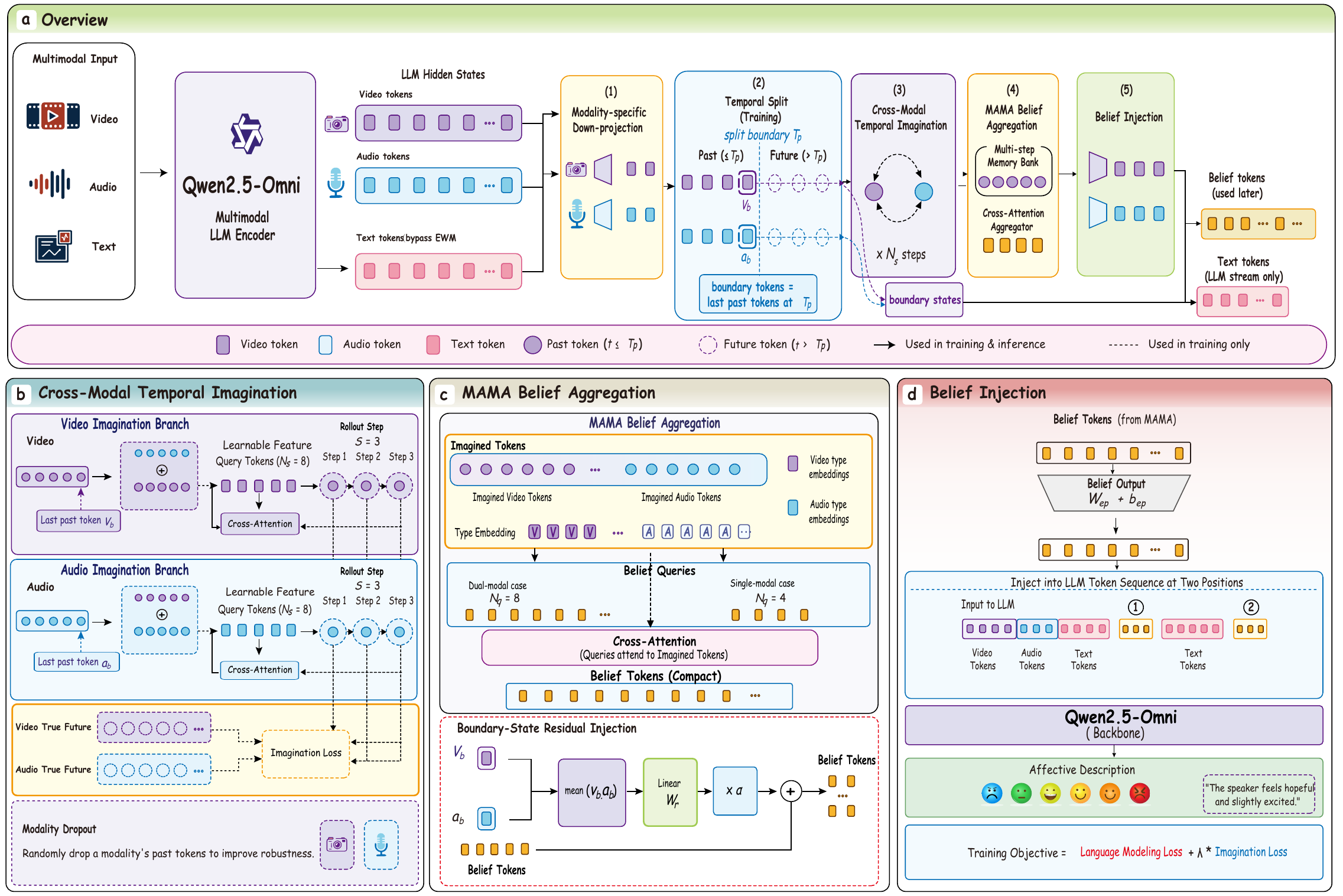}
  \caption{\textbf{Overall framework of AffectVerse.}
\rev{AffectVerse extracts audiovisual hidden states from Qwen2.5-Omni, temporally splits them at $T_p$, imagines future latent tokens with cross-modal multi-step rollout, aggregates imagined tokens through MAMA with boundary tokens $\mathbf{v}_b/\mathbf{a}_b$, and interleaves the resulting belief tokens into the LLM sequence for affective generation.}}
  \label{fig:framework}
  \vspace{-1em}
\end{figure}
\vspace{-1em}
\section{Method}
\label{sec:method}
\vspace{-1em}

We model temporally evolving affect as explicit belief-state updates rather than static classification. Although standard MLLMs provide strong multimodal representations, their hidden states are optimized for immediate responses and do not capture affect dynamics under partial observations.
AffectVerse addresses this with an Emotion World Module (EWM). As shown in Figure~2, EWM comprises Cross-Modal Temporal Imagination, MAMA Belief Aggregation, and Belief Injection.



\vspace{-1em}
\subsection{Multimodal LLM Backbone}
\label{sec:backbone}
\vspace{-0.5em}
We build upon Qwen2.5-Omni~\cite{qwen2025omni}, a multimodal LLM that natively processes video, audio, and text in a unified transformer architecture. The model's \emph{thinker} module tokenizes each modality---video tokens $\{v_1, \ldots, v_M\}$, audio tokens $\{a_1, \ldots, a_N\}$, and text tokens $\{t_1, \ldots, t_K\}$---and processes them jointly to produce hidden states:
\begin{equation}
    \mathbf{H} = \text{Thinker}([\mathbf{v}_{1:M}; \mathbf{a}_{1:N}; \mathbf{t}_{1:K}]) \in \mathbb{R}^{L \times d},
\end{equation}
where $L$ is the total sequence length (including formatting tokens) and $d{=}3584$. We apply LoRA~\cite{hu2022lora} ($r{=}16$, $\alpha{=}32$) to all attention projections (Q, K, V, O) for parameter-efficient fine-tuning, preserving pretrained multimodal capabilities while adapting to emotion-specific reasoning.


\vspace{-0.5em}
\subsection{Emotion World Module}
\label{sec:world_model}
\vspace{-0.5em}
The EWM is the core module of AffectVerse. It is grounded in predictive processing framework from neuroscience~\cite{friston2010free,clark2013whatever}.The brain maintains generative models that continuously predict sensory inputs, and emotions are actively constructed through this predictive inference process~\cite{barrett2017theory}. \rev{As shown in Figure~\ref{fig:framework}, EWM consists of Cross-Modal Temporal Imagination, MAMA Belief Aggregation, and Belief Injection, which respectively predict latent affective continuations, compact them into belief tokens, and insert those tokens into the LLM context.}

\textbf{Preparation: Bottleneck Projection and Temporal Splitting.}
The predictive processing framework holds that the brain must infer hidden states from incomplete sensory evidence~\cite{friston2010free}. We operationalize this by establishing partial observations through temporal splitting. Given hidden states $\mathbf{H}$, we extract video tokens $\mathbf{H}_v \in \mathbb{R}^{M \times d}$ and audio tokens $\mathbf{H}_a \in \mathbb{R}^{N \times d}$, and project them into the working space via modality-specific down-projections:
\begin{equation}
    \mathbf{Z}_v = \mathbf{W}_v^{\downarrow} \mathbf{H}_v \in \mathbb{R}^{M \times d_w}, \quad \mathbf{Z}_a = \mathbf{W}_a^{\downarrow} \mathbf{H}_a \in \mathbb{R}^{N \times d_w}.
\end{equation}
This bottleneck reduces computation while acting as an Information Bottleneck~\cite{tishby2000information}.

Each modality is then split temporally into \emph{past} and \emph{future} portions \rev{at the split boundary $T_{p,m} = \lfloor \kappa \cdot L_m \rfloor$ (shown as $T_p$ in Figure~\ref{fig:framework})}, where $\kappa$ is the keep ratio. \rev{We denote the last retained video and audio boundary tokens as $\mathbf{v}_b=\mathbf{Z}_v[T_{p,v}{-}1]$ and $\mathbf{a}_b=\mathbf{Z}_a[T_{p,a}{-}1]$, respectively.} The future portion $\mathbf{Z}^{\text{fut}}$ serves as a detached self-supervised target for the imagination loss, so the prediction task teaches EWM to infer latent affective continuations from observed prefixes without requiring future tokens at inference. We employ a random keep ratio strategy $\kappa \sim \text{Uniform}[0.7, 1.0)$ during training. During inference, $\kappa{=}1.0$ and all observed audiovisual tokens are retained; EWM still provides additional belief-state context learned from the predictive objective. 

\textbf{Cross-Modal Temporal Imagination.}
A key principle of both world models and predictive processing is \emph{incremental prediction}: the brain does not predict the entire future at once but extends predictions step-by-step~\cite{friston2010free}. In the affective domain, this corresponds to the temporal unfolding of emotional appraisal~\cite{scherer2009emotions}. Furthermore, the brain leverages complementary cross-modal information to improve prediction accuracy.

Each modality has an independent imagination module (unshared parameters), but during prediction each modality attends to both its own and the other modality's past, distinguished by learnable tag embeddings $\mathbf{e}_{\text{self}}, \mathbf{e}_{\text{cross}} \in \mathbb{R}^{d_w}$:
\begin{equation}
    \mathbf{C}_v^{(1)} = [(\mathbf{Z}_v^{\text{past}} + \mathbf{e}_{\text{self}}); \; (\mathbf{Z}_a^{\text{past}} + \mathbf{e}_{\text{cross}})].
\end{equation}
Symmetrically, audio imagination uses $\mathbf{C}_a^{(1)} = [(\mathbf{Z}_a^{\text{past}} + \mathbf{e}_{\text{self}}); (\mathbf{Z}_v^{\text{past}} + \mathbf{e}_{\text{cross}})]$. When only one modality is available, the context reduces to the self-modal past alone.
We do not enforce explicit timestamp-level alignment between audio and video tokens in EWM. Instead, we rely on Qwen2.5-Omni's native multimodal encoding to provide implicit cross-modal temporal correspondence, and perform temporal splitting within each modality stream at the same keep ratio $\kappa$.

We perform $S{=}3$ rollout steps using \rev{$N_s{=}8$ learnable future query tokens per step}. At each step $s$:
\begin{equation}
    \mathbf{F}^{(s)} = \text{CrossAttn}(\mathbf{f}_{1:N_s} + \mathbf{e}_{\text{step}}^{(s)},\; \mathbf{C}_m^{(s)}), \quad \mathbf{C}_m^{(s+1)} = [\mathbf{C}_m^{(1)};\; \hat{\mathbf{Y}}_m^{(s)}],
\end{equation}
where $\hat{\mathbf{Y}}_m^{(s)} = \mathbf{W}_{\text{out}} \mathbf{F}^{(s)}$ is the imagined output appended to the context for the next step, progressively extending the imagination horizon. 
 {Here, $\mathbf{C}_m^{(s)}$ denotes the rollout context for target modality $m \in \{v,a\}$ at step $s$; using $m$ keeps the equation modality-agnostic, with $\mathbf{C}_v^{(s)}$ and $\mathbf{C}_a^{(s)}$ instantiated by the video and audio imagination branches respectively.} All rollout steps share cross-attention parameters, with step embeddings $\mathbf{e}_{\text{step}}^{(s)}$ providing temporal distinction, this encourages a general temporal prediction function. Each step's output is supervised against real future tokens:
\begin{equation}
    \mathcal{L}_{\text{imagine}}^{(s)} = \text{MSE}(\hat{\mathbf{Y}}_m^{(s)}, \bar{\mathbf{Z}}_m^{\text{fut}}) + \tfrac{1}{2}(1 - \text{cos}(\hat{\mathbf{Y}}_m^{(s)}, \bar{\mathbf{Z}}_m^{\text{fut}})),
    \label{eq:imagination_loss}
\end{equation}
where $\bar{\mathbf{Z}}_m^{\text{fut}}$ is obtained by adaptive pooling 
the real future tokens to match \rev{$N_s$}.  {\rev{Concretely, for future sequence $\mathbf{Z}_m^{\text{fut}}\in\mathbb{R}^{L_m^{\text{fut}}\times d_w}$, we apply one-dimensional adaptive average pooling along the temporal axis to produce exactly $N_s$ bins: $\bar{\mathbf{Z}}_m^{\text{fut}}=\text{AdaptiveAvgPool1D}_{N_s}(\mathbf{Z}_m^{\text{fut}})$.} This preserves the coarse temporal ordering of the future segment while matching the number of learnable future queries.} The combined MSE (magnitude alignment) and cosine (directional alignment) losses prevent collapse to mean representations while maintaining semantic consistency. Details are in Appendix~\ref{app:method_details}.

\textbf{MAMA Belief Aggregation.}
Scherer's Component Process Model (CPM)~\cite{scherer2009emotions} posits that the appraisal system independently evaluates different stimulus aspects before integration. MAMA operationalizes this ``appraise independently, integrate later'' principle: the imagination modules (Stage 2) independently produce modality-specific multi-step predictions, each rollout step is supervised by the imagination loss, and \rev{MAMA integrates the multi-step imagined tokens into a unified belief state.}
\rev{The rollout outputs are assembled into a multi-step memory bank with \emph{type embeddings} distinguishing modality identity:}
\begin{equation}
    \mathbf{M} = [\hat{\mathbf{Y}}_v^{(1:S)};\; \hat{\mathbf{Y}}_a^{(1:S)}] + \mathbf{E}_{\text{type}}(\text{type\_ids}) \in \mathbb{R}^{N_M \times d_w},
\end{equation}
where \rev{$N_M = |\mathcal{S}| S N_s$ for available modality set $\mathcal{S}$}. A key design choice is \emph{dynamic belief capacity}: the number of belief query tokens adapts to available modalities \rev{($N_q{=}8$ for dual-modality, $N_q{=}4$ for single-modality)}, with \emph{separate} aggregation modules for each regime. This ensures richer inputs receive proportionally larger representation. Belief queries attend to the memory via cross-attention:
\begin{equation}
    \mathbf{b}^{(\text{out})} = \text{CrossAttn}(\mathbf{b}_{1:N_q},\; \mathbf{M}),
\end{equation}
augmented by a \emph{residual boundary injection} that preserves context at the temporal split point: \rev{$\mathbf{b}^{(\text{final})} = \mathbf{b}^{(\text{out})} + \alpha \cdot \mathbf{W}_{r} \bar{\mathbf{z}}_{\text{bnd}}$}, where \rev{$\bar{\mathbf{z}}_{\text{bnd}}=\text{mean}(\mathbf{v}_b,\mathbf{a}_b)$ is computed from the last past video/audio tokens at $T_p$ (using available modalities only)} and $\alpha$ is a learnable scalar initialized to 0.1.

\textbf{Belief Injection}
\label{sec:belief_injection}
After belief aggregation, the compact belief state is projected back to the LLM's hidden dimension and distributed across the token sequence. The belief tokens are up-projected via \rev{$\mathbf{B}_{\text{inject}} = \text{LayerNorm}(\mathbf{W}_{ep} \mathbf{b}^{(\text{final})} + \mathbf{b}_{ep}) \in \mathbb{R}^{N_q \times d}$}. Rather than inserting all tokens at a single position, we split $\mathbf{B}_{\text{inject}}$ into two halves injected at: (1)~immediately after the last audiovisual token, enriching the text encoding with affective context from the imagined future; and (2)~immediately before the answer region, providing a direct ``emotional prior'' for response generation. This interleaved strategy ensures that the LLM can attend to belief information at multiple stages of reasoning.

During training, audiovisual tokens are truncated according to $\kappa$ by applying a token-level keep mask to the embedded sequence: the removed video/audio positions are dropped before the LM loss is computed, and the attention mask and labels are rebuilt after padding the shortened batch. This partial-observation regularization encourages the LLM to attend to the injected belief state instead of treating it as unused extra context. Belief token labels are set to $-100$ (ignored in loss computation) and their attention mask is set to 1. During inference ($\kappa{=}1.0$), no truncation occurs---all audiovisual tokens are retained alongside the belief tokens, so AffectVerse never discards available evidence at test time. Implementation details are in Appendix~\ref{app:method_details}.
\vspace{-0.5em}
\subsection{Training Objectives}
\label{sec:objectives}
\vspace{-0.5em}
The temporal split is used only for training-time predictive supervision, not as an inference-time input restriction. Detached future audiovisual hidden states supervise EWM to learn affective transitions, while belief-token-augmented partial sequences train the LLM to use predictive belief representations. At inference, $\kappa{=}1.0$ keeps all observed tokens, and belief tokens act as auxiliary dynamic summaries.

AffectVerse is trained end-to-end with two complementary objectives.  {Let $\mathbf{H}_{\text{augmented}}$ denote the LLM hidden-token sequence after interleaved belief injection. Given the original hidden sequence $\mathbf{H}=[\mathbf{h}_1,\ldots,\mathbf{h}_L]$, we split the projected belief tokens as $\mathbf{B}_{\text{inject}}=[\mathbf{B}^{(1)}_{\text{inject}};\mathbf{B}^{(2)}_{\text{inject}}]$ and insert them at the AV--Text boundary $p_{\text{AV}}$ and the Text--Answer boundary $p_{\text{ans}}$.} The \textbf{language modeling loss} is autoregressive cross-entropy on assistant response tokens: 
{
\begin{equation}
    \mathbf{H}_{\text{augmented}}
    =
    [\mathbf{H}_{1:p_{\text{AV}}};
    \mathbf{B}^{(1)}_{\text{inject}};
    \mathbf{H}_{p_{\text{AV}}+1:p_{\text{ans}}-1};
    \mathbf{B}^{(2)}_{\text{inject}};
    \mathbf{H}_{p_{\text{ans}}:L}].
    \label{eq:h_augmented}
\end{equation}
}
\begin{equation}
    \mathcal{L}_{\text{lm}} = -\frac{1}{|\mathcal{T}|} \sum_{i \in \mathcal{T}} \log p(y_i | y_{<i}, \mathbf{H}_{\text{augmented}}).
\end{equation}
where $y_i$ denotes the target token at position $i$ with $y_{<i}$ representing its preceding context, and $\mathcal{T}$ is the set of indices for the tokens being predicted.
The imagination loss(Eq.~\ref{eq:imagination_loss}) is averaged over all $S$ rollout steps and all available modalities $\mathcal{S}$. The total loss balances task performance and temporal prediction:
\begin{equation}
    \mathcal{L}_{\text{total}} = \mathcal{L}_{\text{lm}} + \lambda_{\text{img}} \mathcal{L}_{\text{imagine}}, \quad \lambda_{\text{img}} = 1.0.
    \label{eq:total_loss}
\end{equation}
The $\lambda_{\text{img}} = 1.0$ treats imagination as a co-primary objective alongside language modeling---the imagination loss drives the EWM to produce meaningful belief states, while the language modeling loss ensures task-specific performance.

\vspace{-1em}
\section{Experiments}
\label{sec:experiments}
\vspace{-1em}

We evaluate AffectVerse on nine benchmarks: MER2023/MER2024~\cite{lian2024mer}, MELD~\cite{poria2019meld}, IEMOCAP~\cite{busso2008iemocap}, CMU-MOSI~\cite{zadeh2016mosi}, CMU-MOSEI~\cite{zadeh2018mosei}, CH-SIMS~\cite{yu2020chsims}, CH-SIMS v2~\cite{liu2022chsimsv2}, and OV-MERD+~\cite{lian2024ovmer}. We report weighted F1 for emotion classification and binary accuracy for sentiment analysis, following MER-UniBench~\cite{lian2023affectgpt}; details are in Appendix~\ref{app:datasets}.
\vspace{-1em}
\subsection{Main Results}
\vspace{-1em}
Table~\ref{tab:main_results} presents the main results across nine benchmarks under three modality settings: audio+text (A+T), video+text (V+T), and audio+video+text (A+V+T). ``AffectVerse'' denotes our proposed model.
\begin{table*}[t]
\centering
\scriptsize
\setlength{\tabcolsep}{4pt}
\caption{\textbf{Main results under different modality settings.} 
A/V/T denote audio, video, and text. We report F1 for emotion classification and binary accuracy for sentiment analysis; Mean is averaged over nine benchmarks. 
\textbf{Bold} indicates the best result, and \underline{underline} indicates the second-best result.}
\vspace{-0.5em}
\label{tab:main_results}
\resizebox{\textwidth}{!}{
\begin{tabular}{lccc|cccc|cccc|c|c}
\toprule
\multirow{2}{*}{Model} & \multicolumn{3}{c|}{Modality} & \multicolumn{4}{c|}{Basic} & \multicolumn{4}{c|}{Sentiment} & Fine-grained & \multirow{2}{*}{Mean} \\
 & A & V & T & MER2023 & MER2024 & MELD & IEMOCAP & MOSI & MOSEI & SIMS & SIMS v2 & OV-MERD+ &  \\
\midrule

OneLLM~\cite{han2023onellm} & $\checkmark$ & $\times$ & $\checkmark$ & 25.52 & 17.21 & 28.32 & 33.44 & 64.01 & 54.09 & 63.39 & 61.98 & 22.25 & 41.14 \\
SECap~\cite{xu2023secap} & $\checkmark$ & $\times$ & $\checkmark$ & 40.95 & 52.46 & 25.56 & 36.92 & 55.76 & 54.18 & 59.51 & 57.41 & 36.97 & 46.64 \\
PandaGPT~\cite{su2023pandagpt} & $\checkmark$ & $\times$ & $\checkmark$ & 33.57 & 39.04 & 31.91 & 36.55 & 66.06 & 61.33 & 62.93 & 58.88 & 31.33 & 46.84 \\
Qwen-Audio~\cite{chu2023qwenaudio} & $\checkmark$ & $\times$ & $\checkmark$ & 41.85 & 31.61 & 49.09 & 35.47 & 70.09 & 46.90 & 70.73 & 65.26 & 32.36 & 49.26 \\
Qwen2.5-Omni~\cite{qwen2025omni}& $\checkmark$ & $\times$ & $\checkmark$ & 70.13 & \underline{74.17} & 54.64 & \underline{58.60} & \underline{85.06} & 79.22 & 82.79 & 83.53 & \underline{61.61} & \underline{72.19} \\
SALMONN~\cite{tang2023salmonn} & $\checkmark$ & $\times$ & $\checkmark$ & 55.53 & 45.38 & 45.62 & 46.84 & 81.00 & 67.03 & 68.69 & 65.93 & 45.00 & 57.89 \\
AffectGPT~\cite{lian2023affectgpt}& $\checkmark$ & $\times$ & $\checkmark$ & \underline{72.94} & 73.41 & \underline{56.63} & 55.68 & 83.46 & \underline{80.74} & \underline{82.99} & \underline{83.75} & 59.98 & 72.18 \\
\rowcolor{blue!8}
\textbf{AffectVerse}& $\checkmark$ & $\times$ & $\checkmark$ & \textbf{75.27} & \textbf{76.59} & \textbf{58.32} & \textbf{64.32} & \textbf{85.38} & \textbf{81.25} & \textbf{83.53} & \textbf{84.42} & \textbf{62.87} & \textbf{74.66} \\

\midrule

Otter~\cite{li2023otter} & $\times$ & $\checkmark$ & $\checkmark$ & 16.41 & 14.65 & 22.57 & 29.08 & 52.89 & 50.44 & 57.56 & 53.12 & 16.63 & 34.82 \\
Video-LLaVA~\cite{lin2023videollava} & $\times$ & $\checkmark$ & $\checkmark$ & 36.93 & 30.25 & 30.73 & 38.95 & 56.37 & 61.64 & 53.28 & 57.45 & 34.00 & 44.40 \\
PandaGPT~\cite{su2023pandagpt} & $\times$ & $\checkmark$ & $\checkmark$ & 39.13 & 47.16 & 38.33 & 47.21 & 58.50 & 64.25 & 62.07 & 65.25 & 35.07 & 50.77 \\
Video-ChatGPT~\cite{maaz2023videochatgpt} & $\times$ & $\checkmark$ & $\checkmark$ & 44.86 & 46.80 & 37.33 & 56.83 & 54.42 & 63.12 & 64.82 & 65.80 & 39.80 & 52.64 \\
VideoChat2~\cite{li2023mvbench} & $\times$ & $\checkmark$ & $\checkmark$ & 33.67 & 54.50 & 36.64 & 48.70 & 66.84 & 54.32 & 69.49 & 70.66 & 39.21 & 52.67 \\
LLaMA-VID~\cite{li2023llamavid} & $\times$ & $\checkmark$ & $\checkmark$ & 50.72 & 57.60 & 42.75 & 46.02 & 61.78 & 63.89 & 69.35 & 67.48 & 45.01 & 56.07 \\
VideoChat~\cite{li2023videochat} & $\times$ & $\checkmark$ & $\checkmark$ & 48.73 & 57.30 & 41.11 & 48.38 & 65.13 & 63.61 & 69.52 & 72.14 & 44.52 & 56.71 \\
Chat-UniVi~\cite{jin2023chatunivi} & $\times$ & $\checkmark$ & $\checkmark$ & 57.62 & 65.67 & 45.61 & 52.37 & 54.53 & 63.18 & 68.15 & 66.36 & 48.00 & 57.94 \\
mPLUG-Owl~\cite{xu2024mplug} & $\times$ & $\checkmark$ & $\checkmark$ & 56.86 & 59.89 & 49.11 & 55.54 & 72.40 & 72.91 & 72.13 & 75.00 & 48.18 & 62.45 \\
Qwen2.5-Omni~\cite{qwen2025omni}& $\times$ & $\checkmark$ & $\checkmark$ & 71.14 & 71.97 & 54.48 & 58.27 & 79.02 & 77.18 & 84.79 & 84.13 & 61.49 & 71.39 \\
AffectGPT~\cite{lian2023affectgpt} & $\times$ & $\checkmark$ & $\checkmark$ & \underline{74.58} & \underline{75.29} & \textbf{57.63} & \textbf{62.19} & \underline{82.39} & \underline{81.57} & \textbf{87.20} & \textbf{86.29} & \underline{61.65} & \underline{74.31} \\
\rowcolor{blue!8}
\textbf{AffectVerse}& $\times$ & $\checkmark$ & $\checkmark$ & \textbf{75.28} & \textbf{76.35} & \underline{57.38} & \underline{60.78} & \textbf{82.58} & \textbf{81.78} & \underline{85.76} & \underline{85.00} & \textbf{64.84} & \textbf{74.42} \\

\midrule

PandaGPT~\cite{su2023pandagpt} & $\checkmark$ & $\checkmark$ & $\checkmark$ & 40.21 & 51.89 & 37.88 & 44.04 & 61.92 & 67.61 & 68.38 & 67.23 & 37.12 & 52.92 \\
Emotion-LLaMA~\cite{cheng2024emotionllama} & $\checkmark$ & $\checkmark$ & $\checkmark$ & 59.38 & 73.62 & 46.76 & 55.47 & 66.13 & 67.66 & 78.32 & 77.23 & 52.97 & 64.17 \\
Qwen2.5-Omni~\cite{qwen2025omni}& $\checkmark$ & $\checkmark$ & $\checkmark$ & 77.43 & 77.35 &  55.29 & 60.51 & \underline{83.42} & \underline{81.54} & 85.70 & 85.53 & \underline{63.36} & 74.76 \\
AffectGPT~\cite{lian2023affectgpt} & $\checkmark$ & $\checkmark$ & $\checkmark$ & \underline{78.54} & \underline{78.80} & \underline{55.65} & \underline{60.54} & 81.30 & 80.90 & \textbf{88.49} & \underline{86.18} & 62.52 & \underline{74.77} \\
\rowcolor{blue!8}
\textbf{AffectVerse}& $\checkmark$ & $\checkmark$ & $\checkmark$ & \textbf{80.92} & \textbf{82.39} & \textbf{58.47} & \textbf{63.87} & \textbf{84.45} & \textbf{86.60} & \underline{86.60} & \textbf{86.30} & \textbf{66.50} & \textbf{77.34} \\
\bottomrule
\end{tabular}
}
\vspace{-2em}
\end{table*}
Table~\ref{tab:main_results} shows consistent gains for AffectVerse across modality settings. In full A+V+T, AffectVerse reaches \textbf{77.34 \%} average, exceeding Qwen2.5-Omni (74.76\%) and AffectGPT (74.77\%) by \textbf{+2.57\%}, and ranks first on 8/9 benchmarks. The largest gains appear on CMU-MOSEI (+5.06 over Qwen2.5-Omni; +5.70 over AffectGPT) and MER2024 (+3.59\% over AffectGPT), indicating improved modeling of temporally evolving affect. In dual-modality settings, A+T (74.66\%) is stronger than V+T (74.08\%), consistent with stronger emotional signal in prosody~\cite{barrett2017theory}. On fine-grained OV-MERD+, AffectVerse improves to 66.50\% (+3.98\% over AffectGPT), suggesting that belief-state injection is particularly helpful for subtle categories.  {We follow the MER-UniBench evaluation protocol introduced by AffectGPT, using the same benchmarks, modality settings, and primary metrics.}
 {We defer temporal truncation and missing-modality analysis to the ablation section.} 

\begin{table*}[t]
\centering
\caption{\textbf{Modality/temporal robustness ablation on eight benchmarks.} Three settings: Full (A+V+T), No Audio (V+T), No Video (A+T), evaluated across keep ratios. Left: sentiment benchmarks (Acc-2, \%); Right: classification benchmarks (WA-F1, \%). CMU-MOSI is fully measured.}
\label{tab:abl_all}
\vspace{-2pt}
\begin{minipage}[t]{0.50\textwidth}
\centering
\setlength{\tabcolsep}{3pt}
\scriptsize
\begin{tabular}{@{}ll|ccccc|c@{}}
\toprule
\textbf{Dataset} & \textbf{Setting} & 1.0 & 0.7 & 0.5 & 0.3 & 0.1 & Avg. \\
\midrule
\multirow{3}{*}{CMU-MOSI}
  & Full      & 84.5 & 82.7 & 82.7 & 82.7 & 84.3 & 83.4 \\
  & No Audio  & 82.6 & 81.1 & 81.0 & 81.4 & 81.1 & 81.4 \\
  & No Video  & 85.4 & 86.0 & 84.9 & 86.0 & 84.2 & 85.3 \\
\midrule
\multirow{3}{*}{CMU-MOSEI}
  & Full      & 86.6 & 85.8 & 85.4 & 85.2 & 85.0 & 85.6 \\
  & No Audio  & 80.8 & 80.1 & 79.7 & 79.3 & 79.1 & 79.8 \\
  & No Video  & 81.3 & 80.6 & 80.1 & 79.8 & 79.5 & 80.3 \\
\midrule
\multirow{3}{*}{SIMS}
  & Full      & 86.6 & 84.2 & 83.5 & 83.0 & 82.5 & 84.0 \\
  & No Audio  & 84.8 & 83.4 & 82.8 & 82.3 & 81.9 & 83.0 \\
  & No Video  & 83.5 & 84.8 & 83.2 & 82.7 & 82.3 & 83.3 \\
\midrule
\multirow{3}{*}{SIMS v2}
  & Full      & 86.3 & 85.6 & 85.1 & 84.7 & 84.4 & 85.2 \\
  & No Audio  & 85.0 & 84.3 & 83.9 & 83.5 & 83.1 & 84.0 \\
  & No Video  & 84.4 & 84.4 & 84.0 & 83.6 & 83.2 & 83.9 \\
\midrule
\multirow{3}{*}{\textit{Avg.}}
  & Full      & 86.0 & 84.6 & 84.2 & 83.9 & 84.1 & 84.6 \\
  & No Audio  & 83.3 & 82.2 & 81.9 & 81.6 & 81.3 & 82.1 \\
  & No Video  & 83.7 & 84.0 & 83.1 & 83.0 & 82.3 & 83.2 \\
\bottomrule
\end{tabular}
\end{minipage}%
\hfill
\begin{minipage}[t]{0.50\textwidth}
\centering
\setlength{\tabcolsep}{3pt}
\scriptsize
\begin{tabular}{@{}ll|ccccc|c@{}}
\toprule
\textbf{Dataset} & \textbf{Setting} & 1.0 & 0.7 & 0.5 & 0.3 & 0.1 & Avg. \\
\midrule
\multirow{3}{*}{MER2023}
  & Full      & 80.9 & 80.1 & 79.5 & 79.0 & 78.6 & 79.6 \\
  & No Audio  & 75.3 & 74.9 & 74.6 & 74.3 & 74.0 & 74.6 \\
  & No Video  & 75.3 & 76.0 & 74.9 & 74.6 & 74.2 & 75.0 \\
\midrule
\multirow{3}{*}{MER2024}
  & Full      & 82.4 & 82.9 & 82.0 & 81.5 & 81.1 & 82.0 \\
  & No Audio  & 76.4 & 75.9 & 75.5 & 75.2 & 74.9 & 75.6 \\
  & No Video  & 76.6 & 76.9 & 76.2 & 75.9 & 75.5 & 76.2 \\
\midrule
\multirow{3}{*}{MELD}
  & Full      & 58.5 & 57.9 & 57.5 & 57.2 & 56.8 & 57.6 \\
  & No Audio  & 56.4 & 56.0 & 55.7 & 55.4 & 55.1 & 55.7 \\
  & No Video  & 58.3 & 57.9 & 57.5 & 57.2 & 56.9 & 57.6 \\
\midrule
\multirow{3}{*}{IEMOCAP}
  & Full      & 63.9 & 63.8 & 63.4 & 63.1 & 62.7 & 63.4 \\
  & No Audio  & 60.8 & 60.9 & 60.5 & 60.1 & 59.8 & 60.4 \\
  & No Video  & 64.3 & 60.3 & 59.5 & 58.9 & 58.3 & 60.3 \\
\midrule
\multirow{3}{*}{\textit{Avg.}}
  & Full      & 71.4 & 71.2 & 70.6 & 70.2 & 69.8 & 70.6 \\
  & No Audio  & 67.2 & 66.9 & 66.6 & 66.3 & 66.0 & 66.6 \\
  & No Video  & 68.6 & 67.8 & 67.0 & 66.7 & 66.2 & 67.3 \\
\bottomrule
\end{tabular}
\end{minipage}
\vspace{-1em}
\end{table*}
\vspace{-1em}
\subsection{Ablation Studies}
\vspace{-0.5em}
\textbf{Modality/Temporal Robustness.}
Table~\ref{tab:abl_all} shows graceful degradation under temporal truncation and missing modalities. Full-modality performance remains stable as $\kappa$ decreases; on CMU-MOSI, accuracy stays in [82.65\%, 84.45\%] from $\kappa{=}1.0$ to 0.1, and the no-video setting peaks at $\kappa{=}0.7$ (86.01\%), higher than $\kappa{=}1.0$ (85.38\%). MER2024 exhibits the same trend (82.85\% vs.\ 82.39\%), indicating that moderate truncation can regularize learning instead of hurting it. At $\kappa{=}1.0$, removing audio causes larger drops than removing video on most datasets (e.g., MER2024: $-$6.04\% vs.\ $-$5.80\%), consistent with the dominant role of prosody. At the same time, A+V+T still wins on most benchmarks, confirming cross-modal complementarity. Under missing-modality evaluation, degradation remains limited (MELD no-video: 58.32\% vs.\ 58.47\%), supporting the claim that belief-state modeling improves robustness to partial observations along modality and temporal axes.

\vspace{-0.5em}
\begin{table*}[t]
\centering
\small
\caption{\textbf{Component ablation.} Each row adds one component to the previous configuration. $\Delta$ shows the gain over the LoRA baseline.}
\vspace{-0.5em}
\label{tab:component_ablation}
\setlength{\tabcolsep}{3.5pt}
\begin{tabular}{@{}l|cccccc|c@{}}
\toprule
\textbf{Configuration} & MOSI & MOSEI & MER2024 & MELD & SIMS & IEMOCAP & $\Delta$Avg. \\
\midrule
(a) Qwen2.5-Omni + LoRA & 81.58 & 81.54 & 77.35 & 55.29 & 85.70 & 60.51 & --- \\
(b) + Temporal Splitting & 83.78 & 82.38 & 78.92 & 56.14 & 85.89 & 61.07 & +0.87 \\
(c) + Cross-Modal Imagination & 84.05 & 84.20 & 80.46 & 57.25 & 86.12 & 62.15 & +1.87 \\
(d) + MAMA (single-position inject) & 84.21 & 85.52 & 81.58 & 57.82 & 86.30 & 62.98 & +2.44 \\
(e) + Interleaved Injection (2 pos.) & 84.38 & 86.25 & 82.15 & 58.21 & 86.48 & 63.52 & +2.89 \\
(f) + Residual Boundary Injection & 84.45 & 86.60 & 82.39 & 58.47 & 86.60 & 63.87 & \textbf{+3.07} \\
\bottomrule
\end{tabular}
\vspace{-1em}
\end{table*}

\vspace{-1em}
\subsection{Mechanism Analysis}
\vspace{-1em}
\textbf{Component Ablation.}
Table~\ref{tab:component_ablation} isolates the contribution of each proposed component by incrementally building from a LoRA-tuned Qwen2.5-Omni baseline to the full AffectVerse.
Table~\ref{tab:component_ablation} yields three direct conclusions. First, temporal prediction alone is effective: row (b) improves the baseline by +0.87 average and improves all six benchmarks. Second, cross-modal imagination is the largest single jump (row (c), +1.00\% over row (b)), with strong gains on CMU-MOSEI (+1.82\%) and MER2024 (+1.54\%). Third, belief construction and injection are both necessary: MAMA, interleaved injection, and residual boundary injection add consistent gains from rows (d)--(f).

\begin{table}[t]
\centering
\scriptsize
\captionsetup{skip=2pt}
\caption{\textbf{EWM module ablation.}
$\Delta$ is measured against the LoRA baseline.}
\label{tab:non_predictive_ablation}
\setlength{\tabcolsep}{2.8pt}
\renewcommand{\arraystretch}{1.06}
\resizebox{0.85\linewidth}{!}{
\begin{tabular}{@{}lcccccccc@{}}
\toprule
\textbf{Belief construction} & MOSI & MOSEI & MER24 & MELD & SIMS & IEMO & Avg. & $\Delta$ \\
\midrule
None 
& 81.58 & 81.54 & 77.35 & 55.29 & 85.70 & 60.51 & 73.66 & 0.00 \\
Random tokens 
& 82.10 & 82.20 & 77.80 & 55.70 & 85.85 & 60.90 & 74.09 & $+0.43$ \\
Pooling 
& 83.05 & 83.40 & 79.10 & 56.55 & 86.05 & 61.70 & 74.97 & $+1.31$ \\
\rowcolor{gray!10}
\textbf{EWM} 
& \textbf{84.45} & \textbf{86.60} & \textbf{82.39} & \textbf{58.47} & \textbf{86.60} & \textbf{63.87} & \textbf{77.06} & \textbf{+3.40} \\
\bottomrule
\end{tabular}
}
\vspace{-1.5em}
\end{table}
\textbf{EWM Module Ablation.}
Table~\ref{tab:non_predictive_ablation} ablates the belief construction inside EWM. Random belief tokens
and pooling improve the LoRA baseline, but both remain below the EWM. This indicates that EWM's predictive belief construction contributes beyond added token capacity or simple temporal pooling.

\begin{figure*}[t]
\centering
\includegraphics[width=0.92\textwidth]{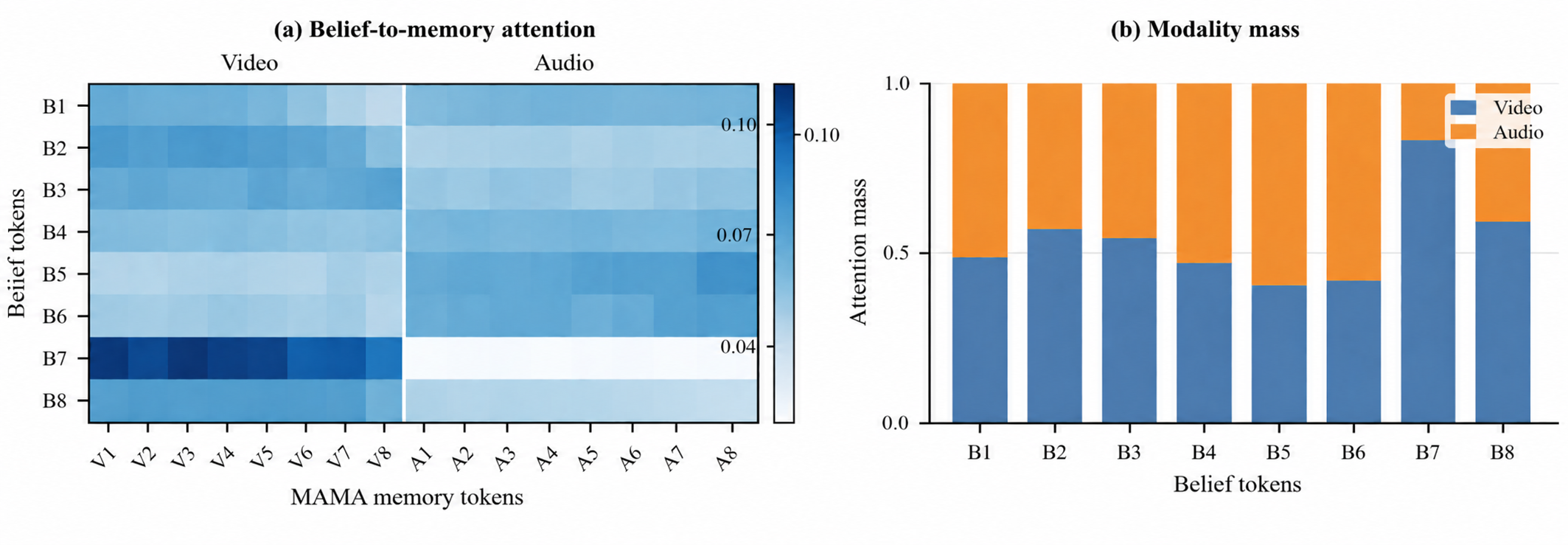}
\vspace{-1em}
\caption{\textbf{Visualization of MAMA belief aggregation.} MAMA belief tokens attend non-uniformly to visual and acoustic memories: some integrate both modalities, while others specialize toward visual (e.g., $B7$) or acoustic evidence (e.g., $B5$--$B6$). This suggests learned modality-aware specialization from type-aware aggregation, rather than manually assigned token roles.}
\label{fig:mama_attention}
\vspace{-2em}
\end{figure*}

\textbf{MAMA Attention Visualization.}
To understand what MAMA learns internally, Figure~\ref{fig:mama_attention} visualizes the cross-attention from the learned belief queries to the final imagined memory tokens on CMU-MOSI validation examples. The left panel shows belief-to-memory attention, where $B1$--$B8$ denote dual-modality belief tokens and $V1$--$V8$ / $A1$--$A8$ denote video and audio memory tokens. The right panel aggregates each row into modality-level attention mass.
Figure~\ref{fig:mama_attention} suggests that MAMA
does more than average multimodal features. Learned belief queries form complementary
\begin{wrapfigure}{r}{0.6\linewidth}
\vspace{-14pt}
\centering
\includegraphics[width=\linewidth]{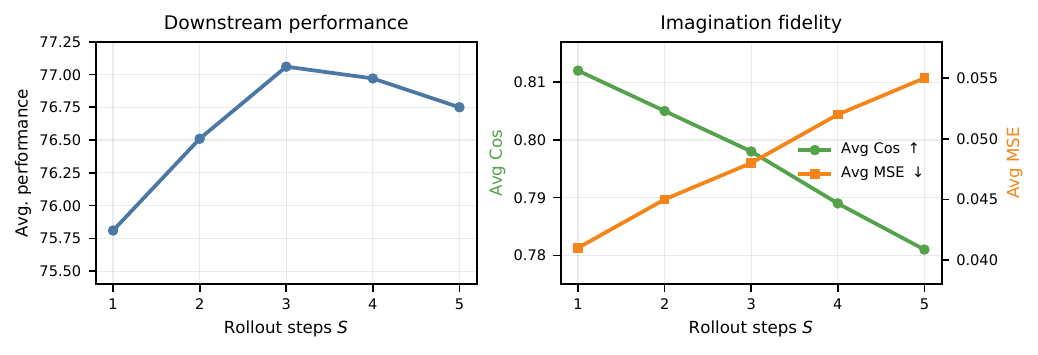}
\vspace{-1.5em}
\caption{\textbf{Rollout depth trade-off.}}
\label{fig:rollout_tradeoff}
\vspace{-1.5em}
\end{wrapfigure}
appraisal slots, either integrating both modalities or focusing on modality-specific evidence. This supports MAMA's goal of building a modality-aware belief state.

\textbf{Imagination Depth: Rollout Steps.} Figure~\ref{fig:rollout_tradeoff} evaluates the effect of rollout depth $S$ and relates it
to imagination fidelity. Performance improves from $S{=}1$ to $S{=}3$ (+1.25\% average), then declines for
$S{\ge}4$ ($-$0.09\% at $S{=}4$, 
$-$0.31\% at $S{=}5$). This indicates that shallow rollouts underfit temporal dynamics, while overly deep rollouts accumulate prediction error. Imagination quality decreases monotonically with depth (cosine 0.812$\to$0.781; MSE 0.041$\to$0.055), while downstream performance peaks at $S{=}3$. Therefore, the best classifier is not the model with the highest one-step fidelity; it is the model 
with the best trade-off between temporal coverage and prediction noise. The asymmetric drift, with a larger relative change in MSE than cosine, also justifies combining magnitude and directional losses.

\begin{wrapfigure}{r}{0.38\linewidth}
\vspace{-1.5em}
\centering
\includegraphics[width=\linewidth]{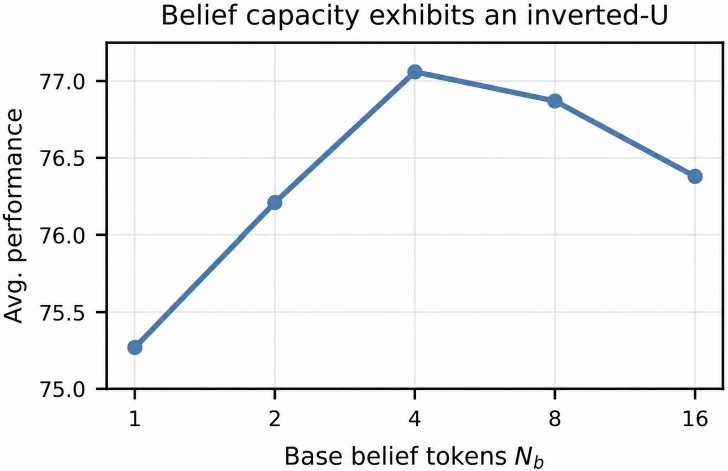}
\vspace{-1em}
\caption{\textbf{Effect of belief token count.}}
\label{fig:belief_capacity}
\vspace{-1.5em}
\end{wrapfigure}

\textbf{Belief Token Capacity.}
Figure~\ref{fig:belief_capacity} studies the impact of the base belief token count \rev{$N_b$}, where 
dual-modality uses \rev{$2N_b$ tokens}, probing how much capacity is needed to encode emotional appraisal. The relation is inverted-U. Too few tokens underfit belief capacity \rev{($N_b{=}1$, $-$1.79\% avg)}, especially 
on CMU-MOSEI and IEMOCAP. Too many tokens dilute signal and increase sequence burden 
\rev{($N_b{=}16$, $-$0.68\% avg)}. The optimum is \rev{$N_b{=}4$}, corresponding to 8 tokens under dual-modality.

\textbf{Cross-Modal vs.\ Self-Modal Imagination.}
\label{sec:ablation_crossmodal}
The cross-modal imagination design is a key differentiator of AffectVerse: rather than predicting each modality's future in isolation, each modality attends to both its own and the other modality's past. Table~\ref{tab:crossmodal_ablation} isolates this contribution by comparing three context modes.
\begin{table*}[t]
\centering
\small
\caption{\textbf{Cross-modal vs.\ self-modal imagination.} Comparison of different imagination context modes. Avg Cos: average imagination cosine similarity.}
\vspace{-0.5em}
\label{tab:crossmodal_ablation}
\setlength{\tabcolsep}{3.5pt}
\begin{tabular}{@{}l|c|cccccc|c@{}}
\toprule
\textbf{Imagination Mode} & Avg Cos $\uparrow$ & MOSI & MOSEI & MER2024 & MELD & SIMS & IEMOCAP & Avg. \\
\midrule
Self-modal only & 0.762 & 83.78 & 82.38 & 78.92 & 56.14 & 85.89 & 61.07 & 74.70 \\
Cross-modal only & 0.741 & 83.52 & 83.15 & 79.68 & 56.82 & 85.95 & 61.85 & 75.16 \\
\rowcolor{gray!10}
\textbf{Full cross-modal} & \textbf{0.798} & \textbf{84.45} & \textbf{86.60} & \textbf{82.39} & \textbf{58.47} & \textbf{86.60} & \textbf{63.87} & \textbf{77.06} \\
\bottomrule
\end{tabular}
\vspace{-1em}
\end{table*}
The results show that full cross-modal imagination outperforms self-modal only by +2.36\% average points, with gains on CMU-MOSEI (+4.22\%) and MER2024 (+3.47\%) where audiovisual complementarity is pronounced. Cross-modal \emph{only} mode (attending exclusively to the other modality) slightly beats self-modal only (+0.46\% avg), indicating that the other modality's temporal history carries predictive
\begin{wrapfigure}{r}{0.65\linewidth}
\vspace{-1em}
\centering
\includegraphics[width=\linewidth]{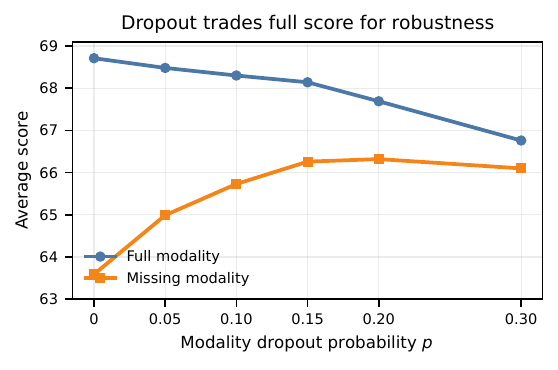}
\caption{\textbf{Effect of modality dropout.}}
\label{fig:dropout_robustness}
\vspace{-1em}
\end{wrapfigure}
value. However, its lower imagination quality (Cos 0.741 vs.\ 0.762) reveals a trade-off: the other modality provides complementary but less direct evidence for predicting a modality's own future. 
The full cross-modal mode combines both signals through learnable tag embeddings ($\mathbf{e}_{\text{self}}, \mathbf{e}_{\text{cross}}$), achieving the highest imagination quality (Cos 0.798) and the best downstream performance. This is consistent with the \emph{maximum likelihood estimation} principle in multisensory integration~\cite{ernst2002humans}, where channels are weighted by reliability.

\textbf{Modality Dropout Ratio.}
\label{sec:ablation_dropout}
Modality dropout ($p{=}0.15$) trains the EWM to form beliefs from partial 
sensory evidence. Figure~\ref{fig:dropout_robustness} reveals the trade-off between full-modality performance and missing-modality robustness within this partial-observation setting.
The trade-off is clear. Without dropout ($p{=}0$), full-modality scores are highest (e.g., IEMOCAP: 
64.32\%), but the missing-modality gap reaches 5.12\%, indicating over-reliance on cross-modal cues that may be unavailable at test time. At $p{=}0.30$, the gap shrinks to 0.66\%, but full-modality performance drops by 1.95\%, suggesting
that the model is trained too often in suboptimal single-modality regimes. At $p{=}0.15$, only 0.57\% are sacrificed in the full setting, while the missing-modality gap is reduced by 63\%, placing it near the Pareto frontier for modality robustness. This pattern is consistent with 
findings in sensory neuroscience that moderate uncertainty promotes robust multisensory integration~\cite{ernst2002humans}.

\textbf{Imagination Loss Weight $\lambda_{\text{img}}$.}
We find $\lambda_{\text{img}}{=}1.0$ gives the best downstream average, confirming that imagination should act as a co-primary objective rather than a weak auxiliary loss.
\begin{table}[t]
\centering
\small
\label{sec:ablation_lambda}
\caption{\textbf{Effect of imagination loss weight $\lambda_{\text{img}}$.} We report Avg Cos (imagination quality) and downstream performance on three representative benchmarks plus the six-benchmark average.}
\label{tab:lambda_ablation}
\setlength{\tabcolsep}{3.5pt}
\resizebox{\textwidth}{!}{
\begin{tabular}{@{}c|c|cccccc|c@{}}
\toprule
$\lambda_{\text{img}}$ & Avg Cos $\uparrow$ & MOSI & MOSEI & MER2024 & MELD & SIMS & IEMOCAP & Avg. \\
\midrule
0 (no img loss) & --- & 83.62 & 82.75 & 78.15 & 55.82 & 85.72 & 60.85 & 74.49 \\
0.01 & 0.725 & 83.85 & 83.92 & 79.48 & 56.58 & 85.95 & 61.62 & 75.23 \\
0.1 & 0.768 & 84.12 & 85.38 & 81.25 & 57.65 & 86.28 & 62.85 & 76.26 \\
0.5 & 0.790 & 84.32 & 86.18 & 82.05 & 58.18 & 86.48 & 63.52 & 76.79 \\
\rowcolor{gray!10}
\textbf{1.0} & \textbf{0.798} & \textbf{84.45} & \textbf{86.60} & \textbf{82.39} & \textbf{58.47} & \textbf{86.60} & \textbf{63.87} & \textbf{77.06} \\
2.0 & 0.805 & 84.28 & 86.35 & 82.12 & 58.25 & 86.42 & 63.58 & 76.83 \\
5.0 & 0.812 & 83.82 & 85.45 & 81.18 & 57.42 & 86.05 & 62.72 & 76.11 \\
\bottomrule
\end{tabular}
}
\vspace{-1em}
\end{table}
Table~\ref{tab:lambda_ablation} shows that the imagination objective is necessary. Without it, EWM degenerates to randomly initialized belief tokens and drops 2.57\% below $\lambda_{\text{img}}{=}1.0$. Increasing $\lambda_{\text{img}}$ improves both imagination 
quality and downstream performance up to 1.0, while larger weights overemphasize 
regression and reduce task performance.

\begin{figure}[htbp]
\centering
\includegraphics[width=\linewidth]{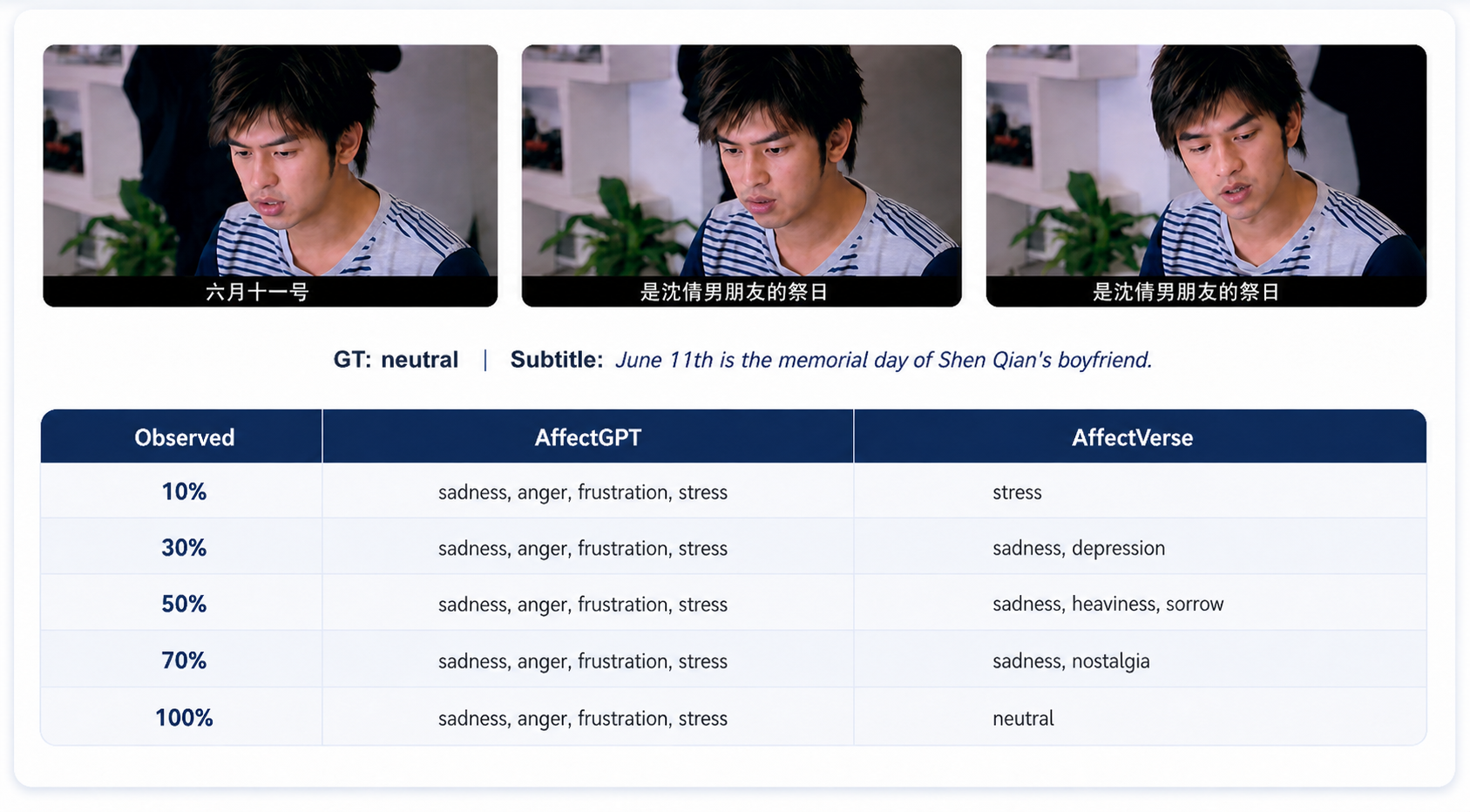}
\caption{\textbf{Case study.} AffectGPT keeps predicting a negative emotion cluster across all observation ratios, while AffectVerse moves from stress under limited observation to the correct neutral prediction when the full audiovisual context is available.}
\label{fig:case_study}
\vspace{-1em}
\end{figure}


\vspace{-1em}
\subsection{Case Study}
\vspace{-0.5em}

In this case study, the text semantics and early facial cues suggest a potentially negative situation, so AffectGPT repeatedly outputs sadness-, anger-, frustration-, and stress-related labels. AffectVerse also predicts stress at 10\% observation, but its belief state becomes more calibrated as additional temporal evidence arrives and finally matches the neutral ground truth at 100\% observation. This example shows that predictive belief-state modeling can separate transient contextual cues from the final affective state.

\vspace{-1em}
\section{Conclusion}
\label{sec:conclusion}
\vspace{-0.5em}
 {\rev{We presented AffectVerse, a Qwen2.5-Omni-based model that augments multimodal emotion recognition with an Emotion World Module composed of Cross-Modal Temporal Imagination, MAMA Belief Aggregation, and Belief Injection.} Rather than treating an affective clip as a static input, AffectVerse predicts latent audiovisual continuations from observed evidence and compresses the resulting transition information into belief tokens for the LLM. Across nine benchmarks, AffectVerse improves over strong MLLM baselines, while ablations show that temporal imagination, cross-modal rollout, MAMA aggregation, and modality-dropout training each contribute to the final performance.}

 {\textbf{Limitations.} AffectVerse is evaluated with fixed rollout and keep-ratio hyperparameters, and its robustness analysis focuses on modality and temporal partial observations rather than broader domain shifts such as noise, accent, topic, or dataset transfer. Extending predictive belief modeling to stronger alignment, domain-shift evaluation, and dialogue-level emotion tracking remains future work.}

\bibliographystyle{plainnat}
\bibliography{references}


\appendix

\section{Detailed Dataset Descriptions}
\label{app:datasets}

We evaluate AffectVerse on nine multimodal emotion and sentiment benchmarks. Below we provide detailed descriptions of each dataset.

\textbf{MER2023}~\cite{lian2024mer}. The Multimodal Emotion Recognition Challenge 2023 dataset contains video clips annotated with 6 discrete emotion categories (angry, happy, neutral, sad, worried, surprise). Each sample consists of a short video segment with corresponding audio and textual transcripts. We use the official training and test splits.

\textbf{MER2024}~\cite{lian2024mer}. The Multimodal Emotion Recognition Challenge 2024 dataset is an updated and extended version of MER2023, featuring the same 6 emotion categories with additional samples and refined annotations. We use the official training set for training and the test set for evaluation.

\textbf{MELD}~\cite{poria2019meld}. The Multimodal EmotionLines Dataset is collected from the TV series \textit{Friends}, containing 7 emotion categories (anger, joy, sadness, neutral, disgust, fear, surprise). Each utterance is associated with video, audio, and textual transcript. We use the official train/dev/test splits.

\textbf{IEMOCAP}~\cite{busso2008iemocap}. The Interactive Emotional Dyadic Motion Capture dataset consists of approximately 12 hours of conversational videos with 4 emotion categories (happy, sad, neutral, angry). We follow the standard 5-fold protocol, using the first 4 sessions for training and session 5 for testing.

\textbf{CMU-MOSI}~\cite{zadeh2016mosi}. The CMU Multimodal Opinion-level Sentiment Intensity dataset contains 2,199 opinion utterance-video segments from YouTube movie reviews, annotated with continuous sentiment scores in $[-3, +3]$. We report binary accuracy (Acc-2) following standard practice.

\textbf{CMU-MOSEI}~\cite{zadeh2018mosei}. The CMU Multimodal Opinion Sentiment and Emotion Intensity dataset is a large-scale extension of CMU-MOSI, containing 23,453 annotated video segments from over 1,000 YouTube speakers. Sentiment annotations are provided as continuous scores. We report binary accuracy on the standard test split (4,659 samples).

\textbf{SIMS}~\cite{yu2020chsims}. The CH-SIMS (Chinese Multimodal Sentiment Intensity) dataset contains 2,281 video clips from Chinese movies, TV series, and variety shows, annotated with multimodal sentiment intensity scores. It provides both unimodal and multimodal annotations. We report binary accuracy on the official test split.

\textbf{SIMS~v2}~\cite{liu2022chsimsv2}. SIMS~v2 is an extended version of CH-SIMS with improved annotation quality and additional samples. It follows the same annotation protocol as SIMS but with a larger and more diverse collection. We report binary accuracy.

\textbf{OV-MERD+}. The Open-Vocabulary Multimodal Emotion Recognition Dataset Plus (OV-MERD+) is a fine-grained emotion benchmark that goes beyond basic categorical labels, requiring models to recognize a broader and more nuanced set of emotional states. We report weighted-average F1-score on the official test split.

All datasets provide aligned video, audio, and text (subtitle/transcript) for each sample. Each input is formatted as a multi-turn conversation following the Qwen2.5-Omni chat template, with the system prompt ``You are an emotion recognition assistant,'' the user message containing video, audio, subtitle context, and the question, and the assistant response containing the emotion label or sentiment description.

\subsection{Evaluation Protocol}

 {We follow the MER-UniBench protocol associated with AffectGPT~\cite{lian2023affectgpt} for dataset coverage, modality settings, and primary metrics. The main table includes externally reported results from this benchmark family as reference points, because rerunning all MLLM baselines under identical training recipes is computationally prohibitive. To avoid overclaiming from heterogeneous training histories, we use these external numbers only for broad benchmark positioning. The method-specific evidence is based on Qwen2.5-Omni and AffectVerse runs in our codebase, together with the controlled ablations that keep the backbone, data format, prompts, and optimization setup fixed.}

\subsection{Evaluation Metrics}

For discrete emotion classification (MER2023, MER2024, IEMOCAP, MELD, and OV-MERD+), we report \textbf{weighted average F1-score (WAF)} as the primary metric, along with overall accuracy when applicable. For sentiment analysis benchmarks (CMU-MOSEI, CMU-MOSI, SIMS, and SIMS~v2), we report \textbf{binary accuracy (Acc-2)} following the benchmark protocol.

\subsection{Implementation Details}

AffectVerse is built upon Qwen2.5-Omni-7B~\cite{qwen2025omni} with LoRA~\cite{hu2022lora} adaptation ($r=16$, $\alpha=32$, dropout $=0.05$) applied to all attention projection matrices (Q, K, V, O). The Emotion World Module operates in a bottleneck working dimension $d_w = 512$, with \rev{$N_s = 8$ future query tokens per rollout step}, $S = 3$ rollout steps using 2-layer cross-attention with 8 heads, and \rev{$N_q = 4$ base belief query tokens (8 for dual-modality)}. The keep ratio $\kappa$ is sampled from $\text{Uniform}[0.7, 1.0)$ during training and fixed at 1.0 during inference. Modality dropout probability is 0.15. The imagination loss weight is $\lambda_{\text{img}} = 1.0$. Videos are sampled at 3 FPS and resized to 224$\times$224. The maximum sequence length is 1024 tokens. The total EWM parameter overhead is $\sim$25M.

Training uses the AdamW optimizer with a cosine-annealed learning rate of $2{\times}10^{-4}$, weight decay $0.01$, and linear warm-up for the first 3\% of steps. The effective batch size is 32 (4 per GPU $\times$ 8 gradient accumulation). We train for 3 epochs on the merged training set of all nine benchmarks. All experiments are conducted on 8 NVIDIA A100-80G GPUs; a full training run takes approximately 18 hours. The LoRA modules and EWM are jointly optimized; the backbone parameters remain frozen.

\subsection{Supplementary Case Study}

\begin{figure}[h]
\centering
\includegraphics[width=0.98\textwidth]{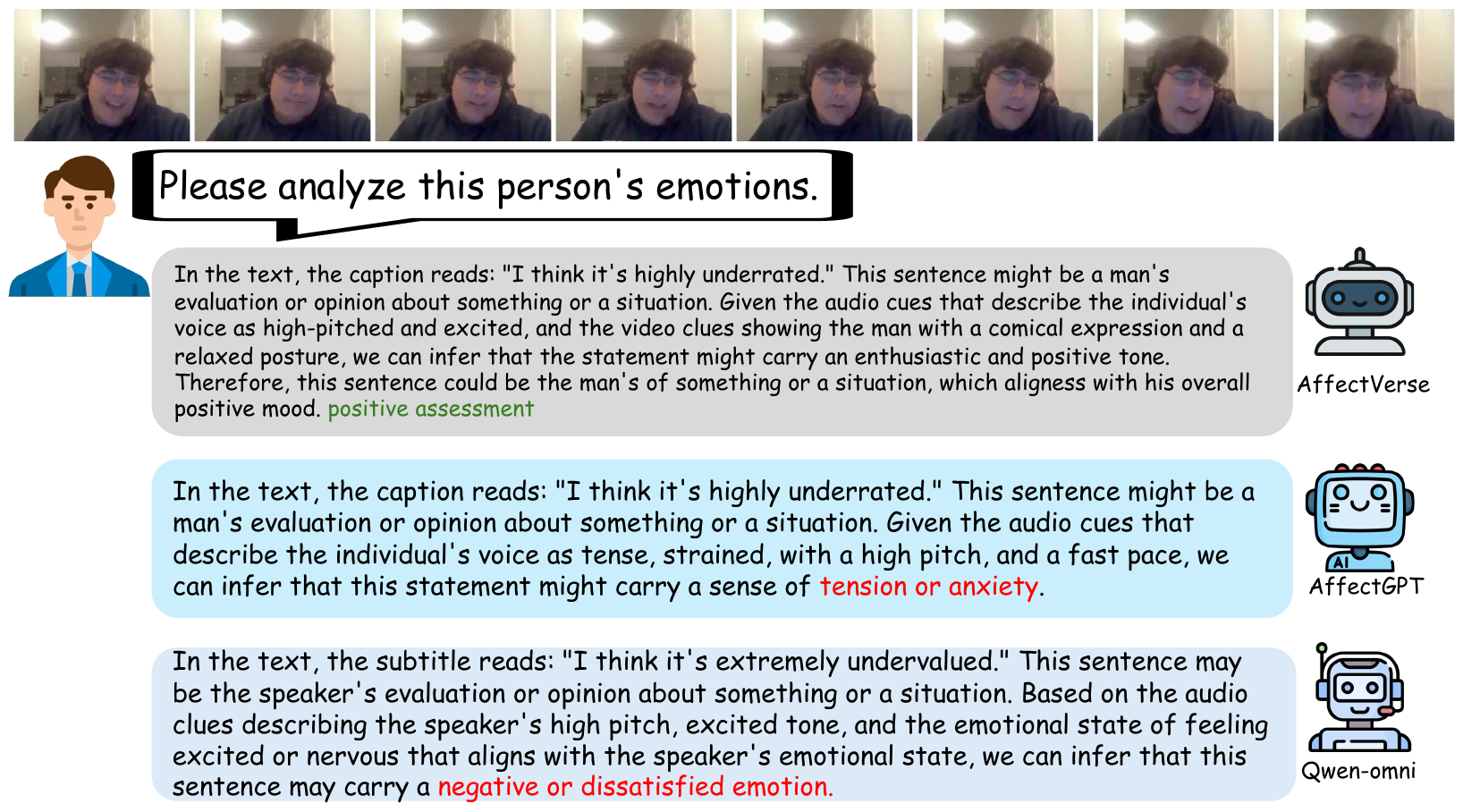}
\caption{\textbf{Supplementary visualization of the full-observation correction case.} This appendix figure provides an alternative compact visualization of the qualitative example in Figure~\ref{fig:case_study}, where AffectVerse corrects from stress under limited observation to neutral under full observation.}
\label{fig:case_study_appendix}
\vspace{-1em}
\end{figure}

Figure~\ref{fig:case_study_appendix} provides an additional view of the qualitative example discussed in Figure~\ref{fig:case_study}. The sample is useful because the subtitle contains a semantically negative event, yet the final ground-truth affect is neutral. Under limited observation, both the model and the baseline can be attracted to negative cues; the distinction is that AffectVerse revises its belief as more temporal evidence becomes available.

The progression from stress to neutral illustrates the intended role of the Emotion World Module: the belief tokens should not merely amplify early emotional cues, but should help the LLM update its interpretation when later audiovisual context contradicts an initially negative reading. This complements the main case study by showing the same correction behavior in a compact appendix format.

\section{Detailed Architecture Specifications}

Table~\ref{tab:arch} summarizes the key architectural hyperparameters of AffectVerse.

\begin{table}[h]
\centering
\small
\caption{Architecture hyperparameters of AffectVerse.}
\label{tab:arch}
\begin{tabular}{lc}
\toprule
\textbf{Component} & \textbf{Configuration} \\
\midrule
\multicolumn{2}{l}{\textit{Backbone (Qwen2.5-Omni-7B)}} \\
Hidden dimension $d$ & 3584 \\
LoRA rank $r$ / scaling $\alpha$ & 16 / 32 \\
LoRA dropout & 0.05 \\
LoRA target modules & Q, K, V, O projections \\
Max sequence length & 1024 \\
\midrule
\multicolumn{2}{l}{\textit{Emotion World Module}} \\
Working dimension $d_w$ & 512 \\
\rev{Future query tokens $N_s$ (per step)} & \rev{8} \\
Imagination cross-attention layers & 2 \\
Attention heads & 8 \\
Rollout steps $S$ & 3 \\
\rev{Belief query tokens $N_q$ (single / dual)} & \rev{4 / 8} \\
Belief aggregation layers & 1 \\
Type embedding classes & 6 \\
Modality tag embeddings & 2 (self / cross) \\
Modality dropout & 0.15 \\
Residual scale $\alpha$ (initial) & 0.1 \\
FFN activation & GELU \\
\midrule
\multicolumn{2}{l}{\textit{Temporal Splitting}} \\
Keep ratio $\kappa$ (training) & $\sim\text{Uniform}[0.7, 1.0)$ \\
Keep ratio $\kappa$ (inference) & 1.0 (no truncation) \\
Injection strategy & Interleaved (2 positions) \\
\midrule
\multicolumn{2}{l}{\textit{Training}} \\
Imagination loss & MSE + 0.5 $\times$ Cosine \\
$\lambda_{\text{img}}$ & 1.0 \\
Total EWM parameters & $\sim$25M \\
\bottomrule
\end{tabular}
\end{table}

\section{Algorithm}

Algorithm~\ref{alg:training} provides a pseudocode summary of the AffectVerse training procedure.

\begin{algorithm}[h]
\caption{AffectVerse Training Step}
\label{alg:training}
\begin{algorithmic}[1]
\REQUIRE Multimodal input $(V, A, T)$, labels $y$
\STATE Tokenize and prepare inputs: $\text{input\_ids}, \text{attn\_mask}, \text{labels}$
\STATE \textbf{Full-observation Thinker Forward:} $\mathbf{H} \leftarrow \text{Thinker}(\text{input\_ids})$ \COMMENT{Used to form detached predictive targets}
\STATE \textbf{Extract modality tokens:} $\mathbf{H}_v, \mathbf{H}_a \leftarrow \text{ExtractAV}(\mathbf{H})$
\STATE \textbf{Sample keep ratio:} $\kappa \sim \text{Uniform}[\kappa_{\min}, 1.0)$
\STATE \textbf{Modality dropout:} randomly drop one modality with $p=0.15$
\STATE \textbf{Bottleneck projection:} $\mathbf{Z}_v \leftarrow \mathbf{W}_v^{\downarrow} \mathbf{H}_v$, $\mathbf{Z}_a \leftarrow \mathbf{W}_a^{\downarrow} \mathbf{H}_a$
\STATE \rev{\textbf{Temporal split:} $\mathbf{Z}_m^{\text{past}}, \mathbf{Z}_m^{\text{fut}}, \mathbf{z}_m^{\text{bnd}}, T_{p,m} \leftarrow \text{Split}(\mathbf{Z}_m, \kappa)$ for $m \in \{v,a\}$}
\STATE \textbf{Cross-modal imagination:} $\hat{\mathbf{Y}}_v^{(1:S)}, \hat{\mathbf{Y}}_a^{(1:S)} \leftarrow \text{Imagine}(\mathbf{Z}^{\text{past}})$ \COMMENT{Multi-step rollout}
\STATE \textbf{Imagination loss:} $\mathcal{L}_{\text{img}} \leftarrow \text{MSE+Cos}(\hat{\mathbf{Y}}^{(1:S)}, \text{stopgrad}(\mathbf{Z}^{\text{fut}}))$
\STATE \rev{\textbf{MAMA:} $\mathbf{b} \leftarrow \text{BeliefAggregation}([\hat{\mathbf{Y}}_v^{(1:S)}; \hat{\mathbf{Y}}_a^{(1:S)}], \text{type\_ids})$ \COMMENT{Dynamic $N_q$}}
\STATE \rev{\textbf{Residual boundary:} $\mathbf{b} \leftarrow \mathbf{b} + \alpha \cdot \mathbf{W}_{r} \bar{\mathbf{z}}_{\text{bnd}}$}
\STATE \rev{\textbf{Up-project:} $\mathbf{B}_{\text{inject}} \leftarrow \text{LN}(\mathbf{W}_{ep} \mathbf{b} + \mathbf{b}_{ep})$}
\STATE \textbf{Partial-observation inject:} $\mathbf{H}_{\text{aug}} \leftarrow \text{Inject}(\mathbf{H}, \mathbf{B}_{\text{inject}}, \kappa)$ \COMMENT{Training only; $\kappa{=}1.0$ at inference}
\STATE $\mathcal{L}_{\text{lm}} \leftarrow \text{CrossEntropy}(\text{LMHead}(\mathbf{H}_{\text{aug}}), \text{labels})$
\STATE $\mathcal{L}_{\text{total}} \leftarrow \mathcal{L}_{\text{lm}} + \lambda_{\text{img}} \mathcal{L}_{\text{img}}$
\STATE Backpropagate $\mathcal{L}_{\text{total}}$ and update parameters
\end{algorithmic}
\end{algorithm}

\section{Detailed Method Formulations}
\label{app:method_details}

This appendix provides extended formulations and theoretical discussions for the Emotion World Module, complementing the concise presentation in \S\ref{sec:world_model}--\S\ref{sec:belief_injection}.

\subsection{Bottleneck Projection and Temporal Splitting}

\textbf{Information Bottleneck Perspective.}
The down-projection $\mathbf{W}_m^{\downarrow} \in \mathbb{R}^{d_w \times d}$ ($d{=}3584 \rightarrow d_w{=}512$) serves a dual purpose: (1)~it reduces computational cost for all subsequent operations, and (2)~it acts as an Information Bottleneck~\cite{tishby2000information} that forces the model to compress multimodal representations, retaining only the most emotionally relevant information.

\textbf{Temporal Splitting Details.} For each modality, we split the projected tokens along the temporal axis:
\begin{equation}
    \mathbf{Z}_v^{\text{past}} = \mathbf{Z}_v[:T_{p,v}], \quad \mathbf{Z}_v^{\text{fut}} = \mathbf{Z}_v[T_{p,v}:], \quad \mathbf{v}_b=\mathbf{Z}_v[T_{p,v}{-}1], \quad T_{p,v} = \lfloor \kappa \cdot M \rfloor.
\end{equation}
\rev{The same splitting applies to audio tokens with boundary $T_{p,a}$ and boundary token $\mathbf{a}_b=\mathbf{Z}_a[T_{p,a}{-}1]$; Figure~\ref{fig:framework} denotes the modality-specific boundary generically as $T_p$.} The future portion $\mathbf{Z}^{\text{fut}}$ serves as a detached self-supervised target---the EWM must learn to imagine these representations from the past alone, but the target itself is used only for training-time predictive supervision.

\textbf{Random Keep Ratio.} We employ a stochastic strategy:
\begin{equation}
    \kappa \sim \begin{cases} \text{Uniform}[\kappa_{\min}, 1.0) & \text{during training} \\ 1.0 & \text{during inference} \end{cases},
\end{equation}
where $\kappa_{\min} = 0.7$. During inference, $\kappa = 1.0$ means no tokens are truncated---the LLM retains all audiovisual tokens while additionally receiving the belief state as an auxiliary predictive summary.

\textbf{Modality Dropout.} We randomly drop an entire modality with probability $p_{\text{drop}} = 0.15$ during training (7.5\% per modality). When a modality is dropped, the EWM switches to single-modality mode, training the model to form emotional beliefs from partial sensory evidence.

\subsection{Cross-Modal Imagination Architecture}

\textbf{Cross-Modal Context.} For video imagination, the context is:
\begin{equation}
    \mathbf{C}_v^{(1)} = [(\mathbf{Z}_v^{\text{past}} + \mathbf{e}_{\text{self}}); \; (\mathbf{Z}_a^{\text{past}} + \mathbf{e}_{\text{cross}})],
\end{equation}
where $\mathbf{e}_{\text{self}}, \mathbf{e}_{\text{cross}} \in \mathbb{R}^{d_w}$ are learnable tag embeddings. Symmetrically, $\mathbf{C}_a^{(1)} = [(\mathbf{Z}_a^{\text{past}} + \mathbf{e}_{\text{self}}); \; (\mathbf{Z}_v^{\text{past}} + \mathbf{e}_{\text{cross}})]$. When only one modality is available, the context reduces to self-modal past alone.

\textbf{Multi-Step Rollout Details.} The complete rollout procedure for each modality $m$:
\begin{align}
    \mathbf{Q}^{(s)} &= \mathbf{f}_{1:N_s} + \mathbf{e}_{\text{step}}^{(s)}, \\
    \mathbf{F}^{(s)} &= \text{CrossAttn}(\mathbf{Q}^{(s)},\; \mathbf{C}_m^{(s)}), \quad s = 1, \ldots, S, \\
    \hat{\mathbf{Y}}_m^{(s)} &= \mathbf{W}_{\text{out}} \mathbf{F}^{(s)}, \quad \mathbf{C}_m^{(s+1)} = [\mathbf{C}_m^{(1)};\; \hat{\mathbf{Y}}_m^{(s)}],
\end{align}
where each cross-attention block consists of 2 layers with Pre-LN, 8 attention heads, and GELU-activated FFN. All rollout steps share the same cross-attention parameters---step embeddings $\mathbf{e}_{\text{step}}^{(s)}$ provide the only temporal distinction, which reduces model size while encouraging a general temporal prediction function.

\subsection{MAMA: Memory and Belief Details}

\textbf{Type Embedding Table.} \rev{MAMA uses the multi-step imagined outputs from each available modality as memory tokens.} The type embedding table $\mathbf{E}_{\text{type}} \in \mathbb{R}^{6 \times d_w}$ maps type IDs to dense vectors (6 types reserved for extensibility):
\begin{equation}
    \text{type\_id}(\mathbf{m}_i) \in \{0: \text{video imagination}, \; 1: \text{audio imagination}\}.
\end{equation}

\textbf{Dynamic Belief Capacity.} The number of belief tokens adapts to available modalities:
\begin{equation}
    N_q = \begin{cases} 2 N_b = 8 & \text{dual-modality} \\ N_b = 4 & \text{single-modality} \end{cases}.
\end{equation}
The dual- and single-modality settings use \emph{separate} belief aggregation modules (independent parameters), allowing each to specialize in its respective information regime.

\textbf{Belief Aggregation.} Learnable belief queries \rev{$\mathbf{b}_{1:N_q} \in \mathbb{R}^{N_q \times d_w}$} attend to all memory tokens through a single cross-attention layer with Pre-LN, 8 heads, and GELU FFN:
\begin{equation}
    \mathbf{b}^{(\text{out})} = \text{CrossAttn}(\mathbf{b}_{1:N_q},\; \mathbf{M}) \in \mathbb{R}^{N_q \times d_w}.
\end{equation}

\subsection{Interleaved Injection Details}

\textbf{Up-Projection.} \rev{$\mathbf{B}_{\text{inject}} = \text{LayerNorm}(\mathbf{W}_{ep} \mathbf{b}^{(\text{final})} + \mathbf{b}_{ep}) \in \mathbb{R}^{N_q \times d}$, where $\mathbf{W}_{ep} \in \mathbb{R}^{d \times d_w}$ and $\mathbf{b}_{ep}\in\mathbb{R}^{d}$.}

\textbf{Injection Positions.} Belief tokens are split into two halves:
\begin{equation}
    \mathbf{B}_{\text{inject}} = [\underbrace{\mathbf{b}_1, \ldots, \mathbf{b}_{N_q/2}}_{\text{Position 1: AV--Text boundary}};\; \underbrace{\mathbf{b}_{N_q/2+1}, \ldots, \mathbf{b}_{N_q}}_{\text{Position 2: Text--Answer boundary}}].
\end{equation}

\textbf{Token-Level Truncation.} During training, a keep mask is built over the embedded token sequence. For each contiguous video/audio region, only the first $\kappa$ fraction is retained and the remaining suffix is removed before computing the LM loss. The shortened examples are then padded to a common length, with attention masks and labels rebuilt to match the new sequence. This trains the LLM to use belief tokens under partial observations; their labels are set to $-100$ (ignored in loss), while their attention mask is set to 1. During inference ($\kappa{=}1.0$), no audiovisual token is removed, and the belief tokens enrich rather than replace the full multimodal context.

\subsection{Training Details}

\textbf{Imagination Loss Averaging.} The imagination loss averages over all rollout steps and available modalities:
\begin{equation}
    \mathcal{L}_{\text{imagine}} = \frac{1}{|\mathcal{S}|} \sum_{m \in \mathcal{S}} \frac{1}{S} \sum_{s=1}^{S} \mathcal{L}_{\text{imagine},m}^{(s)},
\end{equation}
where $\mathcal{S}$ is the set of available modalities after modality dropout. The MSE component enforces magnitude alignment while the cosine component enforces directional (semantic) alignment, preventing collapse to mean representations---consistent with strategies in self-supervised learning~\cite{grill2020byol,caron2021dino}.


\subsection{Per-Step Imagination Quality Breakdown}

Table~\ref{tab:perstep_imagination} provides a detailed breakdown of imagination quality at each individual rollout step. Since all rollout steps share the same cross-attention parameters (with only step embeddings providing temporal distinction), and each step's output is supervised against the same adaptive-pooled future target, we report the cosine similarity between each step's output and the ground-truth future representation.

\begin{table}[h]
\centering
\small
\caption{\textbf{Per-step imagination quality breakdown.} Cosine similarity between each rollout step's imagined output and the real future representation, measured on the validation set. ``---'' indicates that the step does not exist for the given $S$ configuration. Avg Cos and Avg MSE are averaged across all active steps and both modalities.}
\label{tab:perstep_imagination}
\setlength{\tabcolsep}{3pt}
\begin{tabular}{@{}c|ccccc|cc|c@{}}
\toprule
$S$ & Step 1 & Step 2 & Step 3 & Step 4 & Step 5 & Avg Cos $\uparrow$ & Avg MSE $\downarrow$ & Avg Perf $\uparrow$ \\
\midrule
1 & 0.812 & --- & --- & --- & --- & 0.812 & 0.041 & 75.81 \\
2 & 0.811 & 0.799 & --- & --- & --- & 0.805 & 0.045 & 76.51 \\
\rowcolor{gray!10}
\textbf{3} & \textbf{0.810} & \textbf{0.798} & \textbf{0.787} & --- & --- & \textbf{0.798} & \textbf{0.048} & \textbf{77.06} \\
4 & 0.809 & 0.796 & 0.784 & 0.768 & --- & 0.789 & 0.052 & 76.97 \\
5 & 0.808 & 0.795 & 0.783 & 0.766 & 0.751 & 0.781 & 0.055 & 76.75 \\
\bottomrule
\end{tabular}
\end{table}

Several patterns emerge from the per-step breakdown: (1)~Step~1 cosine similarity is stable across all $S$ configurations (0.808--0.812), indicating that the shared-parameter imagination module maintains similar first-step quality regardless of how many subsequent steps are appended. (2)~Each additional step shows a modest quality decrease of ${\sim}$0.012--0.017 in cosine similarity, suggesting that error accumulation is controlled but non-negligible. (3)~The quality gap between the first and last step widens with $S$: from 0.012 ($S{=}2$) to 0.057 ($S{=}5$), which helps explain why excessive rollout steps ($S{\geq}4$) lead to diminishing returns in downstream performance despite providing more imagination tokens. This per-step analysis provides evidence supporting the choice of $S{=}3$ as the best trade-off.


\end{document}